\pdfoutput=1

\documentclass[11pt]{article}

\usepackage{acl}

\usepackage{times}
\usepackage{latexsym}
\usepackage{amsmath}
\usepackage{amssymb}
\usepackage{microtype}

\usepackage[T1]{fontenc}

\usepackage[utf8]{inputenc}

\usepackage{microtype}

\usepackage{graphicx}

\title{Substitution-based Semantic Change Detection using Contextual Embeddings}

\author{Dallas Card \\
  University of Michigan School of Information, Ann Arbor, MI \\
  \texttt{dalc@umich.edu} }

\begin{document}
\maketitle
\begin{abstract}

Measuring semantic change has thus far remained a task where methods using contextual embeddings have struggled to improve upon simpler techniques relying only on static word vectors. Moreover, many of the previously proposed approaches suffer from downsides related to scalability and ease of interpretation. We present a simplified approach to measuring semantic change using contextual embeddings, relying only on the most probable substitutes for masked terms. Not only is this approach directly interpretable, it is also far more efficient in terms of storage, achieves superior average performance across the most frequently cited datasets for this task, and allows for more nuanced investigation of change than is possible with static word vectors.

\end{abstract}

\section{Introduction}

Measuring semantic change is one of the few areas of NLP where contextual embeddings have not yet led to a definitive improvement over previous methods. In particular, the commonly used approach of aligning static embeddings trained on different time periods \citep{hamilton.2016.diachronic} continues to be a surprisingly hard to beat baseline. 

Given that contextual embeddings provide a representation for each occurrence of a word in context, they would seem to be ideally suited to a more nuanced investigation of semantic change. Most attempts to leverage them for this purpose, however, produce quantitatively worse results, while  being less interpretable and requiring more resources.

Here, we present a simplified and improved approach to scalable, interpretable, semantic change detection using contextual embeddings. Inspired by \citet{eyal.2022.large}, we work only with the most probable replacements for masked words, and measure semantic change in terms of the distributions of replacements in each time period. Not only does this better match human judgements, it is highly space efficient, works seamlessly for out-of-vocabulary words, and helps intuitively characterize meaning change and variation.

\section{Background}

Measuring semantic change involves a set of tasks related to determining if and how a term's meaning has changed over time. Here, we focus on the task of measuring the amount of change that has occurred from one time period to another \cite{gulordava.2011.distributional,schlechtweg.2020.semeval}.\footnote{For surveys of computational approaches to lexical semantic change detection, see \citet{kutuzov.2018.diachronic}, \citet{tang.2018.state}, and \citet{tahmasebi.2021.survey}.}

Existing approaches to this task are mostly of two types. The first is associating each term with a single vector per time period and measuring the distance between vectors, of which we take \citet{hamilton.2016.diachronic} to be representative. As a variation on this, several authors have proposed averaging the output of  contextual embedding models to get a single vector per term in each time period, but this has generally not led to an improvement over using static vectors \citep{martinc.2020.leveraging,kurtyigit.2021.lexical,liu.2021.statistically}. A related approach is to represent words in terms of their nearest neighbors using static word vectors \citep{hamilton.2016.cultural,gonen.2020.simple}, but this does not show a clear improvement over other static embedding methods \citep{montariol.2021.scalable}.

A second type of approach begins with various methods for word sense induction, then measures change in terms of the relative prevalence of a term's different senses \citep{frermann.2016.bayesian,hu.2019.diachronic,arefyev.2020.bos,arefyev.2021.interpretable}. In some cases, authors simply cluster contextual representations for each term, and measure differences in the distributions of clusters between two time periods, rather than dealing with explicit word senses \citep{giulianelli.2020.analysing,martinc.2020.capturing,montariol.2021.scalable}. 

Despite the additional information  provided by contextual embedding models, methods using type embeddings (as opposed to token), continue to be competitive. For example, on the recent SemEval multilingual semantic change detection task, none of the top four systems used token embeddings \citep{schlechtweg.2020.semeval}. Methods using contextual embeddings have done better on some more recent mono-lingual shared tasks \citep{kutuzov.2021.rushifteval,zamora.2022.lscdiscovery}, but have not yet been evaluated with a consistent setup across multiple languages.

\section{Methods}

Building on \citet{eyal.2022.large}, we represent each token in the corpus (or a sufficiently large sample of them) by a small set of probable replacement terms from a contextual embedding model. However, whereas \citet{eyal.2022.large} did this for the purpose of word sense disambiguation, we do so for the purpose of measuring semantic change.

For each sampled occurrence of each term, we mask the term of interest, feed the masked context
through a model, and obtain the predicted token probabilities corresponding to the mask token.\footnote{Words that get tokenized into multiple word pieces are replaced by a single mask token.} From these, we save only the   
top-$k$ most probable words (excluding stopwords and partial word pieces), and discard the rest.

For a given term in a particular time period, we then count how many times each word in the model vocabulary has appeared as a top-$k$ replacement for that term, and normalize this by its sum, giving us a distribution over replacements.
To obtain a raw score of semantic change between two time periods, we compute the Jensen-Shannon Divergence (JSD) between the two distributions representing the same term in different time periods. However, as we show below, the raw JSD scores are strongly correlated with term frequency. Thus, to obtain a scaled metric, we convert the raw JSD scores into a quantile, comparing the raw score for a term of interest to other terms with similar frequency.

Compared to saving the full output vector per token, this approach only requires a miniscule amount of storage per token, and thus does not require the kind of heuristic dropping of tokens employed by \citet{montariol.2021.scalable}. In addition, the dominant meanings of a word in each context can be summarized by the terms which occur most frequently among the top-$k$ replacements. Although such replacements are limited to the terms which exist in the model vocabulary, in practice this is sufficient to represent a nuanced set of meanings, and works even for words which get tokenized into multiple word pieces, as we show below.

More formally, given two corpora C1 and C2,
let the count of token $v$ as a top-$k$ replacement for term $t$ in corpus $c$ be:
\begin{equation}
\textrm{count}(v, t, c) = \Sigma_{i=1}^{N_c(t)} \mathbb{I}[v \in R(t, i, k)],
\end{equation}
where $R(t, i, k)$ is the set of top-$k$ most probable replacements for occurrence $i$ of term $t$ (excluding stopwords and partial word pieces in the model vocabulary), and $N_c(t)$ is the number of sampled occurrence of term $t$ in corpus $c$.\footnote{Unlike \citet{eyal.2022.large}, we do not combine probabilities for different forms of the same lemmas in the model vocabulary. In addition, we do not exclude the target term from the top-$k$ replacements, except implicitly for terms which get split into multiple word pieces.}

Let $\Delta_t^c$ by the distribution of top-$k$ replacement counts for term $t$ in corpus $c$, obtained by dividing the corresponding vector of counts (i.e., [$\textrm{count}(\cdot, t, c)$]) by the sum over the model vocabulary.
The raw change score for term $t$ is given by the JSD between the two distributions:
\begin{equation}
\textrm{raw}(t) = \textrm{JSD}\left( \Delta_t^{C1}, \Delta_t^{C2} \right).
\end{equation}

Finally, we correct for frequency effects by rescaling the raw JSD scores against the scores for terms  with similar frequency as the target term, giving us a quantile scaled in [0, 1]:
\begin{equation}
\textrm{scaled}(t) = \Sigma_{s \in T(t)} \mathbb{I}[\textrm{raw}(t) \geq \textrm{raw}(s)]  / | T(t) |,
\end{equation}
where $T(t)$ is the set of terms with similar frequency to term $t$ (excluding term $t$ itself). More specifically, we compare against all terms within a fixed factor of the target frequency:
\begin{equation}
T(t) = \{s: \textrm{fr}(t)/F \leq \textrm{fr}(s) \leq \textrm{fr}(t) \times F, s \ne t\},
\end{equation}
where $\textrm{fr}(t)$ is the frequency of term $t$ in the corpus, with window factor $F$.

\section{Experiments}

To evaluate our method we make use of datasets for which there have been prior evaluations of methods across multiple languages, 
following standards established by past work for the sake of a head-to-head comparison.\footnote{Code to replicate these experiments is available at  \url{https://github.com/dallascard/SBSCD}}

\subsection{Data}

We use five datasets with words labeled in terms of semantic change between two time periods. Four of these are from SemEval 2020 Task 1: Unsupervised Lexical Semantic Change Detection (SE;  \citealp{schlechtweg.2020.semeval}). These datasets contain 31 to 48 terms from four languages, graded in terms of change by human raters, along with accompanying corpora to be used in estimating the amount of change. The fifth dataset (GEMS) comes from \citet{gulordava.2011.distributional}, and contains 100 words labeled in terms of semantic change from the 1960s to 1990s. As with most recent papers which use this dataset, we use the Corpus of Historical American English (COHA; \citealp{coha}) for measuring change in the GEMS words.

\subsection{Experimental Details}

For each dataset, we fine tune an appropriate BERT model to the union of the two associated unlabeled corpora using continued masked language model training with the HuggingFace \texttt{transformers} package. We then index the corpora to find all occurrences of each word. For all target words, along with a random set of 10,000 background terms, we randomly sample up to 4,000 occurrences of each from the associated corpora. We process all sampled tokens as described above to obtain and store the top-$k$ replacements for each, with $k=5$. Using the replacements obtained from the model, we compute raw JSD scores for each term. Finally, we convert these to scaled scores by comparing to the background terms that have frequency within a factor of two of the target term (i.e., $F=2$).

Following past work, we evaluate using Spearman correlation with human ratings, comparing against the best results from recent papers. In particular, we include two results based on slight variations on \citet{hamilton.2016.diachronic}, one of which was the best performing method in the SemEval competition \citep{pomsl.2020.circe}, as well as methods using contextual embeddings \citep{martinc.2020.capturing,montariol.2021.scalable}. For fully experimental details, please refer to Appendix \ref{sec:appendix_experiments}.

\subsection{Results}

Full results are given in Table \ref{tab:results}. Although our method is not uniformly better than all previous methods on all dataset, it does produce the best result on average, as well as  improvements on GEMS, SE English and SE Latin. 

\begin{table*}[]
    \small
    \centering
    \begin{tabular}{l r r r r r r r}
         & GEMS & SE Eng & SE Ger & SE Lat & SE Swe & Average & Average (weighted) \\
        \hline \hline
        Number of words & 96$^*$ & 37 & 40 & 48 & 31 & \\
        \hline \hline
        \emph{Static Embedding Methods} \\
        \hline
        \citet{pomsl.2020.circe} & - & 0.422 & \bf{0.725} & 0.412 & 0.547 & - & - \\        
        \citet{montariol.2021.scalable} [static] &  0.347 & 0.321 & 0.712 & 0.372 & \bf{0.631} & 0.477 & 0.452 \\                
        \hline \hline
        \emph{Contextual Embedding Methods} \\
        \hline
        \citet{martinc.2020.capturing} & 0.510 & 0.313 & 0.436 & 0.467 & -0.026 & 0.340 & 0.394
        \\        
        \citet{montariol.2021.scalable} [contextual] & 0.352 & 0.437 & 0.561 & 0.488 & 0.321 & 0.432 & 0.422 \\
        Scaled JSD & \bf{0.535} & \bf{0.547} & 0.563 & \bf{0.533} & 0.310 & \bf{0.498} & \bf{0.514} \\
    \end{tabular}
    \caption{Spearman correlation results on five datasets, including both an unweighted average and an average weighted by number of words. \citet{pomsl.2020.circe} was the best submission from SemEval 2022 Task 1, but did not evaluate on GEMS. \citet{montariol.2021.scalable} included results using static vectors, as well as several variations on their own method using contextual embeddings, of which we take the one with the highest average performance. \citet{martinc.2020.capturing} only evaluated on GEMS, so we report the replication results from
    \citet{montariol.2021.scalable}. $^*$We exclude four terms from GEMS to match past work; for full results on GEMS, please refer to Appendix \ref{sec:appendix_gems}.}
    \label{tab:results}
\end{table*}

\begin{figure}
    \centering
    \includegraphics[width=\columnwidth]{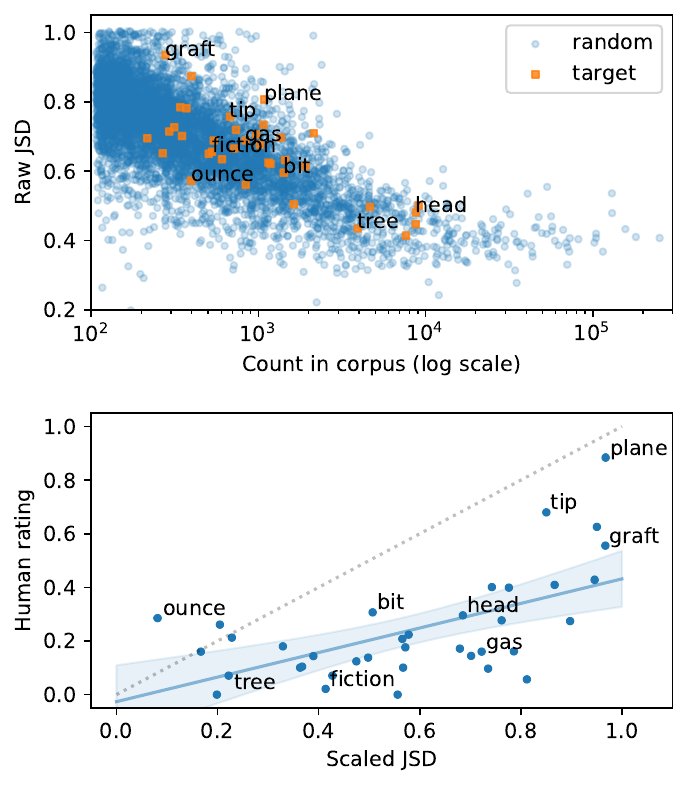}
    \caption{Top: Raw JSD scores for both target and randomly chosen background terms in the SE English dataset, plotted against term counts. Bottom: Human ratings for SE English, plotting against scaled JSD scores, along with a fitted regression line (solid) and the 1:1 diagonal (dotted). Select terms in Table \ref{tab:examples} are labeled.}
    \label{fig:raw_scores}
\end{figure}

As an example to better understand these results, the raw JSD scores from our method are shown in Figure \ref{fig:raw_scores} (top) for the SE English data, with select terms labeled. As can be seen, there is a strong relationship between term frequency and raw JSD, hence the need to rescale the raw scores relative to terms with similar frequency. After rescaling, we see a strong correlation between our final semantic change scores and the human ratings, as shown in Figure \ref{fig:raw_scores}  (bottom) for the SE English data.

\begin{table*}[]
    \small
    \centering
    \begin{tabular}{p{0.55cm}  p{0.75cm} p{0.45cm} p{0.6cm} p{1.15cm} p{4.5cm} p{4.9cm}}
    Word &  SE rating & SE rank & Scaled JSD & Scaled JSD rank & Corpus A substitutes (1810–1860) & Corpus B substitutes (1960–2010) \\
    \hline \hline
    plane & 0.88 & 1 & 0.97 & 1 & plane line planes point surface lines & plane aircraft planes jet airplane car \\
    graft & 0.55 & 4 & 0.97 & 2 & tree plant stock vine fruit wood & corruption bribery fraud crime violence \\           
    tip & 0.68 & 2 & 0.85 & 7 & tipped tip covered end filled tips give & tip tips end tipped edge point top ends \\    
    gas & 0.16 & 23 & 0.72 & 14 & gas gases vapor air fire water & gas  gasoline oil gases fuel water air \\
    head & 0.30 & 10 & 0.68 & 16 & head face hand heads hands eyes & head face heads hand body hands eyes \\    
    bit & 0.31 & 9 & 0.51 & 23 & bit piece sort little pieces bits kind & bit little lot touch tad piece bits pieces \\
    fiction & 0.02 & 35 & 0.41 & 27 & fiction history literature art poetry & fiction fact fantasy story stories novels \\
    tree & 0.07 & 33 & 0.22 & 33 & trees tree plants branches plant wood & trees tree plants woods branches bushes \\
    ounce & 0.28 & 11 & 0.08 & 37 & ounce inch pounds hour acre dollars & ounce pounds inch inches cups pieces \\    
    \end{tabular}
    \caption{Example terms from the SE English dataset, showing the most common substitutes from our approach.}
    \label{tab:examples}
\end{table*}

As with the approach of \citet{hamilton.2016.diachronic}, our method supports direct interpretation of semantic change. To understand the change in a word's typical usage, we can look at the overall most common replacements from each time period. Table \ref{tab:examples} shows the scores and rankings of several selected terms from SE English, along with the most common substitutes from each time period. 

Looking at the results, we can see, for example, strong agreement with human annotators on a dramatic change in the meaning of \emph{plane} (comparing 1810--1860 vs.~1960--2010), from the geometric concept to the flying machine. On the other hand, our results suggest that human raters may have slightly underestimated the amount of change in the meaning of \emph{graft}, which was previously used mostly in reference to vegetation, but now most commonly refers to corruption.\footnote{Note that because \emph{graft} is not a term in the BERT vocabulary, the term itself does not appear as a potential substitute, but the results remain interpretable nonetheless.} 

By contrast, \emph{ounce} may be a case where our method has underestimated the change that has taken place. Older usages seem to map more generically to a wider range of quantities (hence the appearance among the early substitutes of \emph{hour},  \emph{acre}, and \emph{dollars}), whereas modern usage seems more restricted. Indeed, we do find some difference in the distribution of substitutes between the two time periods, but less of a difference than is typical for words with similar frequency, hence the low final score from our method (see Figure \ref{fig:raw_scores}).

Although we do not emphasize it in this paper, of our method can easily be combined with the approach of \citet{eyal.2022.large} to further investigate meaning changes, by inferring senses from the term replacements, and looking at how their usage varies by time period. In particular, for each target term, we can construct a graph from the set of term substitutes (as nodes), where edge weights represent the number of top-$k$ clusters in which two substitutes co-occur. Following \citet{eyal.2022.large}, we experiment with Louvain community detection to identify sense clusters from these graphs for each term of interest, and use Jaccard similarity to associate each mention with a sense cluster, based on substitute overlap (see Appendix \ref{sec:appendix_experiments} for details). 

Inspecting the distribution of these senses over time helps to distinguish the gradual adoption of existing senses from the creation of new ones. 
For example, the most common sense of \emph{plane} is captured by the sense cluster \{\emph{aircraft}, \emph{jet}, \emph{airplane}, \emph{car}\}, and as expected, this sense is not found in the 1810--1860 English data, except for two instances which appear to be errors in the inferred sense. By contrast, the second most common sense---\{\emph{planes}, \emph{line}, \emph{point}, \emph{surface}\}---appears in both time periods, but is much more common in the earlier time.

This approach also provides more insight into how the meaning of \emph{graft} has changed. The most common sense cluster is the horticultural meaning \{\emph{tree}, \emph{plant}, \emph{stock}, \emph{vine}\}, and this meaning occurs in both time periods, but is much more common in the earlier one. A second cluster, corresponding to illicit activity---\{\emph{corruption}, \emph{violence}, \emph{bribery}, \emph{fraud}\}---occurs only in the later time period. This clustering method also surfaces a third sense with a medical meaning---\{\emph{transplant}, \emph{surgery}, \emph{disease}, \emph{drug}\}---which is not revealed by the top few overall most common replacements given in Table \ref{tab:examples}.

\section{Discussion and Related Work}

As noted by others, new and larger datasets for rigorously evaluating semantic change are badly needed \citep{tahmasebi.2021.survey}. Existing datasets are relatively small, and are mostly based on inspecting a limited number of examples per term. Unfortunately, determining ground truth for semantic change is challenging, and producing such resources is costly. Ideally, future datasets for evaluation should be larger, both to allow for more robust evaluation, and to have sufficient targets for both hyperparameter tuning and evaluation.

In addition to the dataset we have used in this paper, two others are available from shared tasks on Spanish and Russian, respectively \citep{kutuzov.2021.rushifteval,zamora.2022.lscdiscovery}. Both of these are comparable in size to the GEMS dataset used here. Unfortunately, they are less useful for evaluation because most submissions to these shared tasks only evaluated on the task data, and not on other datasets. As shown by the replication of \citet{martinc.2020.capturing} in \citet{montariol.2021.scalable}, a method can sometimes perform well on one language but fail to generalize to others. As such, we have based our evaluation on datasets for which there has been a consistent evaluation of methods across multiple languages. As future work, a careful replication study of all methods from each competition on all available datasets, including an assessment of sensitivity to hyperparameters, would be highly informative.

Besides \citet{eyal.2022.large}, The closest prior work to ours is \citet{kudisov.2022.bos}, who use dynamic patterns to generate many variations on example usages sampled from the given corpora. These variations are then used to generate hundreds of replacement terms from a masked language model with associated probabilities. These probabilities are averaged (heuristically combining replacements with differing numbers of word pieces) to obtain a mean vector for each sampled instance. Finally, semantic change is computed as the average cosine distance between all pairs of vectors across corpora. This method was evaluated as part of the LSCDiscovery shared task on Spanish \citep{zamora.2022.lscdiscovery}. Preliminary work on this method was described in \citet{arefyev.2021.interpretable}, where a slightly different version of it was evaluated on the RuShiftEval shared task on Russian \citep{kutuzov.2021.rushifteval}.

Compared to \citet{kudisov.2022.bos}, our approach is considerably simpler, and better suited to storing representations of a complete corpus for subsequent analysis and exploration. In particular, we only consider a small number of substitutes for each example (storing only the top-$k$ most probable terms, without the associated probabilities). We do not use dynamic patterns, and only consider terms in the model vocabulary as potential substitutes. We also associate each term with a single distribution over the model vocabulary per time period (not per mention), and use Jensen-Shannon divergence to more naturally measure the distance between distributions. Importantly, we also correct for frequency effects, as described above.

Although our approach avoids the onerous storage requirements of methods which save full contextual vectors, it still requires considerable processing time to obtain the top-$k$ replacements for all tokens. Future work could explore smaller or more efficient models for this purpose.\footnote{See Appendix \ref{sec:appendix_alt_models} for results using various model sizes.}

Finally, despite its simplicity, measuring the cosine distance between aligned static vectors remains a strong and efficient baseline \citep{hamilton.2016.diachronic}. More work is needed to determine where contextual embeddings can offer sufficient advantage in measuring semantic change to justify their greater computational cost.

Compared to static embeddings, our approach is weakest on the German and Swedish datasets, which could relate to the quality of the pretrained models that are available for those languages, the data used for pretraining, or perhaps issues that arise in tokenization of the reference corpora. For a tentative exploration of some possible factors, please refer to Appendix \ref{sec:appendix_lang_diffs}.

\section{Conclusion}

We have presented a simplified and improved approach to measuring semantic change using contextual embeddings, based on the Jensen-Shannon Divergence between the distributions of the most probable replacements for masked tokens in different time periods, corrected for frequency effects. This approach achieves superior performance on average, while remaining directly interpretable, with vastly reduced storage requirements. 

\section*{Limitations}

There are several limitations to this work which should be kept in mind. First and foremost, the datasets for evaluating the measurement of semantic change are relatively small, meaning that any estimates of correlation with human judgements will be relatively high variance. In addition, although the SemEval data includes text from four languages, there is no guarantee that these methods will work as well as they do on other languages or other time periods. Moreover, our approach depends on the use of pretrained language models, and the quality (or existence) of these and other relevant resources will vary by language. 

In addition, like all methods, our approach involves numerous small choices, such as the number of background terms to sample, the number of samples taken, and the value of $k$ in choosing top substitutes. We have kept our choices for these consistent across all five datasets, and these values have not been tuned. As such, different choices could result in better or worse correlation with human judgements. It is also worth noting that the human judgements collected by the creators of these datasets may involve errors or noise. It is possible that a different sample of data, or having different people evaluate the same data, would produce different judgements. 

For exploring the variation in word meanings, we have used the approach of \citet{eyal.2022.large} directly, with the only differences being that we mask terms of interest (allowing us to work with terms that do not exist in the model vocabulary), and do not combine multiple forms of lemmas when getting the top-$k$ terms. We adopt this approach because it is especially easy to combine with our own work, but  different methods for word sense induction might lead to different conclusions about the different meanings of a term that existed in any particular time period. In addition, any conclusions drawn are necessarily limited to the corpora that are used, most of which will be a highly biased sample of all text that was produced by all people for any given period of time.

\section*{Ethical Considerations}

This work only uses well established datasets for the purposes for which they were designed (studying changes in languages and evaluating measurement of semantic change), thus poses few  ethical concerns that did not already exist for these data. Nevertheless, it is worth emphasizing that all of  methods discussed in this paper only return, at best, a noisy estimate of semantic change. Words are used differently by different people, and attempts to measure changes in language inevitably simplify the diversity of uses into a single number, which discards a great deal of nuance. As such, any work applying these methods to measure semantic change should be aware of their limitations and proceed carefully.

\section*{Acknowledgements}

Many thanks to Kaitlyn Zhou and anonymous reviewers for helpful comments and suggestions.

\bibliography{anthology,custom}
\bibliographystyle{acl_natbib}

\clearpage

\appendix

\section{Experimental Details}
\label{sec:appendix_experiments}

For each dataset, we use a BERT model, preferring a high quality monolingual model where available.
For GEMS and SE English, we use \texttt{bert-large-uncased}. For SE Latin we use \texttt{bert-base-multilingual-uncased}, \texttt{deepset/gbert-large} for SE German, and \texttt{KB/bert-base-swedish-cased} for SE Swedish,  with all models available through HuggingFace.
In all cases, we first adapt the model to the dataset by doing continued masked language model training for five epochs on the union of the two associated corpora.

For the SemEval data, the corpora are provided in both raw and lemmatized formats, with the target terms given as lemmas. Because the contextual embedding models have been trained on non-lemmatized text, we prefer to embed mentions using the raw (non-lemmatized data). However, because of uncertainty about how the raw text was lemmatized, we begin by aligning the lemmatized data to the non-lemmatized text. We then index terms in the lemmatized data (for both target terms and random background terms), and then map these indices to indices in the corresponding non-lemmatized data, which we then sample to get replacements.

To do the alignment, we begin by tokenizing the text, and then removing the punctuation from both the lemmatized and non-lemmatized text, storing indices to allow mapping back to the original token sequences in the non-lemmatized data. For each pair of texts (a raw and a lemmatized form), we first identify tokens that occur exactly once in each, and align the positions of these to each other, as long as the ordering of these tokens is consistent. We then recursively do this for the subsequences between each adjacent pair of aligned tokens. Given these landmark alignments, (using exact matches), we then attempt to align all remaining substrings between each pair of aligned tokens, (adding padding tokens as necessary), using Levenshtein distance as a heuristic way to evaluate possible token alignments. Finally, we do a post-alignment correction to consider inserting a padding token in each position to correct for occasional off-by-one errors, and taking the best scoring overall alignment. 

By inspecting the target tokens in the raw (non-lemmatized text) that are obtained using this alignment (based on indexing target terms in the lemmatized version, then mapping these indices to the non-lemmatized text using the alignment), we find that the vast majority of mentions are properly aligned. To eliminate the small number of alignment errors, we only keep tokens that are at least two characters in length where the non-lemmatized form comprises at least 0.02\% of the total number of indexed terms for a given lemma, and where the first letter of the indexed token matches the first letter of the target lemma. To account for a small number of special cases (such as examples in SE Latin where a word sometimes starts with ``j'' and sometimes with ``i'', (presumably due to OCR errors), we create a handful of exceptions to the first letter rule. For full details of this alignment process and exceptions, please refer to replication code.\footnote{\url{https://github.com/dallascard/SBSCD}}

In addition, for the SE English data, target terms (only) are given with specific part of speech tags. However, to better match a random sample of background lemmas, we ignore part of speech in our experiments, and index all occurrences of each target term in the lemmatized data. Future work could explore the impact of restricting measurements to certain parts of speech, both for target and background terms.

For GEMS, where the targets are not lemmatized, we ignore lemmatization and simply sample from all exact matches of the target terms as tokens in the raw text. As with past work, we combine the multiple annotations for the GEMS data by averaging their scores.

All masked tokens are fed into the appropriate model with up to 50 tokens to either side from the original context, which returns a probability distribution over the model vocabulary.
When computing the top-$k$ most probable substitutes, we follow \citet{eyal.2022.large} and exclude  stopwords and partial word pieces (i.e., those that start with \texttt{\#\#}). For GEMS and SE English, we use the stopword list from the Snowball stemmer.\footnote{\url{http://snowball.tartarus.org/algorithms/english/stop.txt}} For SE Latin, we use a Latin stopword list from the Perseus Digital Library.\footnote{\url{https://www.perseus.tufts.edu/hopper/stopwords}} For SE German and SE Swedish, we use the respective stopword lists from NLTK.\footnote{\url{https://www.nltk.org/}}

For the exploration of sense clusters in the main paper using Louvain community detection, we use the same data as used in measuring semantic change, keeping $k=5$, but we exclude the target term itself when gathering the top-$k$ substitutes.\footnote{In practice, this is done by initially saving the top-($k+1$) substitutes, and dropping the target term for the purpose of clustering, where necessary.} We then construct a weighted graph for each target term, where nodes represent substitutes, and edge weights correspond to the number of top-$k$ replacement sets in which each pair of replacements appear together.

To obtain sense clusters, we use the implementation of Louvain community detection in \texttt{networkx} with default parameter settings, to detect clusters in the graph.\footnote{\url{https://networkx.org/}} Finally, we associate each instance of a target term with a corresponding cluster using Jaccard similarity between the instance's set of top-$k$ replacements and the terms in the cluster.

All of these experiments were run on either an NVidia RTX A6000 or A5000 GPU.

\section{Alternative Models}
\label{sec:appendix_alt_models}

In order to investigate the effect of model size on the performance of our approach to measuring semantic change, we try a range of model sizes for BERT on the English datasets, all available from HuggingFace. The results are shown in Table \ref{tab:alt_eng_models}. As can be seen, there is a clear correlation between model size and task performance for the SE English data, but this is not the case for the GEMS dataset, perhaps because the COHA corpora used for GEMS provides longer contexts for term mentions (see Appendix \ref{sec:appendix_lang_diffs}). 

\begin{table*}[h!]
    \centering
    \small
    \begin{tabular}{l r r}
        Model & GEMS & SE English \\
        \hline
        \texttt{google/bert\_uncased\_L-4\_H-256\_A-4} (mini) & 0.559  &  0.433 \\
        \texttt{google/bert\_uncased\_L-4\_H-512\_A-8} (small) & 0.544 & 0.495 \\
        \texttt{google/bert\_uncased\_L-8\_H-512\_A-8} (medium) & 0.538 & 0.522 \\
        \texttt{google/bert\_uncased\_L-12\_H-768\_A-12} (base) & 0.541 & 0.512\\      
        \texttt{bert-base-uncased} & 0.509 &  0.525 \\
        \texttt{bert-large-uncased} & 0.535 & 0.547 \\        
    \end{tabular}
    \caption{Results on the English datasets (Spearman correlation) using a range of BERT model sizes on HuggingFace.}
    \label{tab:alt_eng_models}
\end{table*}

We also demonstrate the effect of using a multilingual model, rather than a language specific model, for all datasets other than SE Latin (for which we are already using a multilingual model in the main paper). As can be seen in Table \ref{tab:multiling_models}, the multilingual model uniformly results in worse performance, demonstrating the importance of having a strong language-specific model for measuring semantic change in this way.

\begin{table*}[t]
    \centering
    \small
    \begin{tabular}{l r r r r}
        Model & GEMS & SE Eng & SE Ger & SE Swe  \\
        \hline
        \texttt{bert-base-multilingual-uncased} & 0.524 & 0.480 & 0.481 & 0.209 \\
        Language specific model (from Table \ref{tab:results} in main paper) & 0.535 & 0.547 &  0.563 & 0.310 \\
    \end{tabular}
    \caption{Results when using a multilingual model, compared to the language specific models used in the paper.}
    \label{tab:multiling_models}
\end{table*}

\section{Exploring Performance Differences Across Languages}
\label{sec:appendix_lang_diffs}

Using the method presented in the main paper, our results were better than using static word vectors for English and Latin, but worse for German and Swedish. Unfortunately, we do not yet have a satisfactory explanation for this discrepancy in performance. Notably, other approaches using contextual embeddings (e.g., \citealp{montariol.2021.scalable}), have also performed worse on these languages (relative to approaches based on \citealp{hamilton.2016.diachronic}).

Several possible explanations suggest themselves for why methods based on contextual embeddings might struggle. For example, tokenization used for these models breaks some words into multiple word pieces, which is not an issue for static embeddings. Another consideration is the amount of context in which the examples occur in the reference corpora (since static vectors typically only use very small context windows, whereas contextual embedding models are capable of using much longer contexts). We might also consider factors relevant to all methods, such as the number of examples given for each target term, or the number of different word forms in which each lemma occurs in the corpora provided. 

Although several of these factors perhaps help to explain why performance on English is especially good (relative to static vectors), they do not provide a convincing way to explain the differences in performance observed on the other languages. In particular, the SE English data has the highest proportion of target words that occur in the model vocabulary (without being broken into multiple word pieces), and these lemmas occur in text using the fewest number of surface forms per target.

By contrast, the other languages tend to have more surface forms, on average, with fewer of the target terms occurring in the corresponding model vocabulary, but Swedish is mid-range on the later (with German being lowest). Latin, by contrast, tends to have more examples of target terms per corpus in both time periods (with German again the lowest), but Swedish is between English and Latin. The Swedish model does have a larger vocabulary, but it is not as large as the multilingual model we used for Latin. Quantitative summaries of these factors are presented for reference in Table \ref{tab:language_factors}.

Ultimately, perhaps the best explanation has to do with the quality of the underlying pretrained models available for each language. Given that different models for different languages were trained on entirely different data, this seems like a highly relevant source of potential differences. Unfortunately, is it difficult to assess the overall quality of pretrained models across languages, so all of these explanations essentially remain no more than hypotheses for further investigation.

\begin{table*}[h!]
    \small
    \centering
    \begin{tabular}{l l r r r r r}
        Dataset & Model & \begin{tabular}{@{}r@{}}Median lower \\ target count \end{tabular} & \begin{tabular}{@{}r@{}}Median \\ target \\ forms\end{tabular} & \begin{tabular}{@{}r@{}} Median \\ context \\ length \end{tabular} & \begin{tabular}{@{}r@{}}\% targets \\ as whole \\ words\end{tabular} & \begin{tabular}{@{}r@{}}Vocab \\ size\end{tabular}  \\
        \hline
        GEMS & \texttt{bert-large-uncased} & 93 & 1 & 191 & 97.0 & 30522  \\
        SE Eng & \texttt{bert-large-uncased} & 209 & 4 & 26 & 95.6 & 30522 \\
        SE Ger & \texttt{deepset/gbert-large} & 101 & 7 & 28 & 22.9 & 31102 \\
        SE Lat & \texttt{bert-base-multilingual-uncased} & 472 & 8 & 28 & 25.0 & 105879 \\
        SE Swe & \texttt{KB/bert-base-swedish-cased} & 249 & 9 & 25 & 74.2  & 50325 \\
    \end{tabular}
    \caption{Quantitative summary statistics of various factors which we might be expected to affect differences in performance across languages (relative to approaches based on static word embeddings). Median lower target count is the median across target terms of the number of examples of each target term in the corpus with the lower count (early or later). Median target forms is the median across examples of the number of surface forms corresponding to each target lemma. Median context length is the median number of tokens in which target terms occur. \% targets as whole words is the percent of target terms which exist in the model vocabulary. Vocab size is the number of words in the model vocabulary. Ultimately, none of these provides a convincing explanation for observed differences.}
    \label{tab:language_factors}
\end{table*}

\section{Additional Results on GEMS}
\label{sec:appendix_gems}

The GEMS dataset has been used for evaluation by many additional papers, beyond those discussed in the main body of this paper. However, these have not all used consistent metrics and corpora, making comparison difficult. For completeness, we include additional results here, as shown in Table \ref{tab:gems}.

The GEMS dataset was originally introduced by \citet{gulordava.2011.distributional}, from whom we obtained the labeled data.
These authors reported results in terms of Pearson correlation, and used multiple datasets for measuring semantic change, including the Google Books Corpus.  \citet{frermann.2016.bayesian} also used this dataset for evaluation, but used different additional data (beyond COHA), and reported results in terms of Spearman correlation.

More recent papers using this dataset (from \citealp{giulianelli.2020.analysing} onwards) have tended to make use of the COHA data from the 1960s and 1990s as the corpus in which to measure change, to correspond to the periods used in the annotation process, which we also use for our results in this paper. \citet{martinc.2020.capturing} reported very strong results on this dataset, but subsequent work from the same authors \citep{montariol.2021.scalable} revealed that this method performed relatively poorly on the SemEval datasets, as reported in Table \ref{tab:results} in the main paper.

\begin{table}[h!]
    \small
    \centering
    \begin{tabular}{l r r}
        Paper & Pearson & Spearman \\    
        \hline \\        \citet{gulordava.2011.distributional} & 0.386 & -  \\
        \citet{frermann.2016.bayesian} & - & 0.377\\
        \citet{giulianelli.2020.analysing} [99] & 0.231 & 0.293 \\           
        \citet{martinc.2020.capturing} [96] & 0.560 & 0.510 \\ \citet{montariol.2021.scalable} [96] & - & 0.352 \\
        Scaled JSD [96] & 0.532 & 0.535 \\
        Scaled JSD [99] & 0.541 & 0.553 \\
    \end{tabular}
    \caption{Additional results on the GEMS dataset from \citet{gulordava.2011.distributional}. Note that not all papers reporting results on this dataset used the same corpora or evaluation metric, hence we report both Pearson and Spearman correlation, and restrict ourselves to the COHA dataset, which was used by all authors. Numbers in brackets show the number of target terms excluded. We evaluate using the exclusions of both \citet{giulianelli.2020.analysing} [99] and \citet{martinc.2020.capturing} [96] to enable a full comparison. Note that the high correlation reported on this dataset by  \citet{martinc.2020.capturing}  did not seem to transfer to the SemEval datasets, as shown by \citet{montariol.2021.scalable} and Table \ref{tab:results} in the main paper.}
   \label{tab:gems}
\end{table}

Different authors have excluded different numbers of words from the 100 target terms in evaluation. \citet{giulianelli.2020.analysing} excluded \emph{extracellular} due to insufficient occurrences in COHA during the 1960 and 1990s, which we also exclude for the same reason. \citet{martinc.2020.capturing} and \citet{montariol.2021.scalable} excluded \emph{assay}, \emph{extracellular}, \emph{mediaeval}, and \emph{sulphate} because they were split into multiple tokens by BERT. Because we mask the target terms, multi-piece words are not a problem, but for completeness we evaluate using the exclusions of both \citet{giulianelli.2020.analysing} and \citet{martinc.2020.capturing} and report both in Table \ref{tab:gems}.

\end{document}